\documentclass{article}
\usepackage{ijcai21}

\usepackage{times}
\usepackage{soul}
\usepackage{url}
\usepackage[hidelinks]{hyperref}
\usepackage[utf8]{inputenc}
\usepackage[small]{caption}
\usepackage{graphicx}
\usepackage{amsmath}
\usepackage{amsthm}
\usepackage{xspace}
\usepackage{booktabs}
\usepackage{algorithm}
\usepackage{algorithmic}
\usepackage{bm}
\usepackage{multirow}
\newcommand{\tabincell}[2]{\begin{tabular}{@{}#1@{}}#2\end{tabular}}
\newcommand{\citet}[1]{\citeauthor{#1}~\shortcite{#1}}
\newcommand{\eg}{\emph{e.g.,}\xspace}
\newcommand{\ie}{\emph{i.e.,}\xspace}
\newcommand{\paratitle}[1]{\vspace{1.5ex}\noindent\textbf{#1}}
\newcommand\ignore[1]{}

\makeatletter
\def\@fnsymbol#1{\ensuremath{\ifcase#1\or \dagger\or *\or \ddagger\or
		\mathsection\or \mathparagraph\or \|\or **\or \dagger\dagger
		\or \ddagger\ddagger \else\@ctrerr\fi}}
\makeatother

\title{Pretrained Language Models for Text Generation: A Survey}

\author{
	Junyi Li$^{1,3}$\footnote{Equal contribution.} \and
	Tianyi Tang$^{2}$\footnotemark[1] \and
	Wayne Xin Zhao$^{1,3}$\footnote{Corresponding author.} \And
	Ji-Rong Wen$^{1,2,3}$ \\
	\affiliations
	$^{1}$Gaoling School of Artificial Intelligence, Renmin University of China \\
	$^{2}$School of Information, Renmin University of China \\
	$^{3}$Beijing Key Laboratory of Big Data Management and Analysis Methods \\
	\emails
	\{lijunyi,steven\_tang,jrwen\}@ruc.edu.cn, batmanfly@gmail.com
}

\begin{document}

\maketitle

\begin{abstract}
Text generation has become one of the most important yet challenging tasks in natural language processing (NLP). The resurgence of deep learning has greatly advanced this field by neural generation models, especially the paradigm of pretrained language models (PLMs). In this paper, we present an overview of the major advances achieved  in the topic of PLMs for text generation. As the preliminaries, we present the general task definition and briefly describe the mainstream architectures of PLMs for text generation. As the core content, we discuss how to adapt existing PLMs to model different input data and satisfy special properties in the generated text. We further summarize several important fine-tuning strategies for text generation. Finally, we present several future directions and conclude this paper. Our survey aims to provide text generation researchers a synthesis and pointer to related research.
\end{abstract}

\section{Introduction}

Text generation, which is often formally referred as natural language generation, has become one of the most important yet challenging tasks in natural language processing (NLP). It aims to produce plausible and readable text in human language from input data (\eg a sequence and keywords). Researchers have developed numerous techniques for a wide range of applications of text generation~\cite{li2021textbox}. For example, machine translation generates text in a different language based on the source text~\cite{YangHHHJ20}; summarization generates an abridged version of the source text to include salient information~\cite{guan2020survey}.

With the recent resurgence of deep learning, various works have been proposed to solve text generation tasks based on recurrent neural networks (RNN)~\cite{li2019generating}, convolutional neural networks (CNN)~\cite{GehringAGYD17}, graph neural networks (GNN)~\cite{li2020knowledge}, and attention mechanism~\cite{BahdanauCB14}. One of the advantages of these neural models is that they enable  end-to-end learning of semantic mappings from input to output in text generation. Besides,  neural models are able to learn low-dimensional, dense vectors to implicitly represent linguistic features of text, which is also useful to alleviate data sparsity.


Despite the success of neural models for text generation, a major performance bottleneck lies in the availability of large-scale  datasets. Existing datasets for most of supervised text generation tasks are rather small (except machine translation). Deep neural networks usually have a large number of parameters to learn, which are likely to  overfit on these small datasets and do not generalize well in practice. 


In recent years, the paradigm of pretrained language models (PLMs) is thriving~\cite{elmo}. The idea is to first pretrain the models in large-scale corpus and then fine-tune these models in various downstream tasks to achieve state-of-the-art results. It is widely recognized that PLMs can encode a large amount of linguistic knowledge from corpus and induce universal representations of language. Therefore, PLMs are generally beneficial for downstream tasks and can avoid training a new model from scratch~\cite{gpt3}. Moreover, with the increasing of computational power and the emergence of Transformer architecture~\cite{transformer}, PLMs have advanced from shallow to deep and achieved outstanding performance in many tasks, such as BERT~\cite{bert} and GPT~\cite{gpt2}. Therefore, researchers have proposed various methods to solve text generation tasks based on PLMs. Pretrained on large-scale corpus, PLMs are able to understand natural language accurately and express in human language fluently, both of which are critical abilities to fulfill the text generation tasks. Existing surveys in this area have only partially reviewed some related topics. \citet{ZaibSZ20} and \citet{guan2020survey} provided a synthesis to the research on some text generation subtasks, \ie dialogue systems and summarization, but did not go broader to the other important generation tasks. \citet{abs-2003-08271} summarized two generations of PLMs for the whole NLP domain and introduced various extensions and adaption approaches of PLMs. To the best of our knowledge, our survey is the first work that presents a comprehensive review of PLMs for text generation. It aims to provide text generation researchers a synthesis and pointer to related research. 

To start with, we first present a general task definition with the formulations of different text generation tasks in Section~\ref{sec-back}, and then briefly describe the mainstream architectures of PLMs that are used in text generation in Section~\ref{sec-model}.   
Since the core of text generation is to model the semantic mappings from input to output, we further organize the major advances with respect to the two aspects of  \emph{input} and \emph{output} in Section~\ref{sec-in}-\ref{sec-out}. For input, we mainly discuss how to adapt existing PLMs to different data types. For output, we study how to satisfy special properties for the generated text. 
Furthermore, we summarize several important fine-tuning strategies for text generation in Section~\ref{sec-finetune}. Finally, we present several future directions and conclude this paper in Section~\ref{sec-future}. 


\section{Task and Typical Applications}
\label{sec-back}
In what follows, we formally define the text generation task. The core of text generation is to generate a sequence of discrete tokens $\mathcal{Y}=\langle y_1,\dots,y_j,\dots,y_n \rangle$, where each $y_j$ is drawn from a word vocabulary $\mathcal{V}$. In most cases,  text generation is conditioned on input data, such as attributes, text and structured data, which is denoted as $\mathcal{X}$. Formally, the text generation task can be described as:

\begin{equation}\label{eq-formula}
	P(\mathcal{Y}|\mathcal{X})=P(y_1,\dots,y_j,\dots,y_n|\mathcal{X}).
\end{equation}

According to input $\mathcal{X}$, we next introduce several typical applications of text generation:

\textbullet~If $\mathcal{X}$ is not provided or a random noise vector $\bm{z}$, this task will degenerate into language modeling or unconditional generation task~\cite{gpt2}, which aims to generate text without any constraint.

\textbullet~If $\mathcal{X}$ is a set of discrete attributes (\eg topic words, sentiment labels), the task becomes topic-to-text generation or attribute-based generation~\cite{ctrl}. The information in $\mathcal{X}$ plays the role of guiding  the text generation process and controlling the modes of the generated text.

\textbullet~If $\mathcal{X}$ is structured data like knowledge graph or table, this task will be considered as KG-to-text or table-to-text generation, called data-to-text generation~\cite{li2021planning}. This task aims to generate descriptive text about structured data.

\textbullet~If $\mathcal{X}$ is multimedia input such as image and speech, the task becomes image caption \cite{abs-2003-01473} or speech recognition \cite{fan2019unsupervised}. The core of image caption is to generate a description of an image, while speech recognition enables programs to process human speech into a text format.

\textbullet~The most common form of $\mathcal{X}$ is also a text sequence, and there exist several applications such as machine translation, summarization and dialogue system. Machine translation~\cite{ConneauL19} aims to translate  text from one language into another language automatically,  summarization~\cite{ZhangWZ19} is focused on generating condensed summary of a long document, and dialogue system~\cite{abs-1901-08149} is designed to converse with humans using natural language.

\begin{table}[t]
	\centering
	\begin{tabular}{c|c}
		\toprule[1pt]
		Input $\mathcal{X}$                            & Tasks                         \\
		\hline
		 Random noise                         &Unconditional text generation \\		
		\hline
		\multirow{2.5}{*}{Discrete attributes}      & Topic-to-text generation \\ \cline{2-2}
		                                        & Attribute-based generation \\
		\hline
		Structured data       & Data-to-text generation                      \\
		\hline
  		\multirow{2.5}{*}{Multimedia}      & Image caption \\ \cline{2-2}
  		                                       & Speech recognition  \\
		\hline
		\multirow{3.8}{*}{Text sequence}           & Machine translation                \\ \cline{2-2}
		                 &                        Summarization              \\ \cline{2-2}
		               &                         Dialogue system            \\
		\bottomrule[1pt]
	\end{tabular}
	\caption{Major tasks and inputs for text generation.}
	\label{tab:text}
\end{table}

We present the formulations for the major text generations in Table~\ref{tab:text}.

\section{Standard Architectures for Text Generation}
\label{sec-model}
Pretrained language models (PLMs) are pretrained with a mass of unlabelled text data and can be fine-tuned on downstream generation tasks. Pretrained on large-scale corpus, PLMs encode massive linguistic and world knowledge into vast amounts of parameters, which can enhance the understanding of language and improve the generation quality. The idea of pretraining is inspired by human beings, \ie we transfer and reuse our old knowledge of what we have learned in the past to understand new knowledge and handle a variety of new tasks. In this way, PLMs can successfully perform on new tasks with their old experience and knowledge.

Owing to the great achievements that Transformer~\cite{transformer} has made, almost all PLMs employ the backbone of Transformer. For the text generation tasks, some of PLMs utilize the standard Transformer architecture following basic encoder-decoder framework, while the others apply a decoder-only Transformer. Next, we will introduce these two methods successively.

\paragraph{Encoder-decoder Transformer.} A standard Transformer utilizes the encoder-decoder architecture, which is composed of two stacks of Transformer blocks. The encoder is fed with an input sequence, while the decoder aims to generate the output sequence based on encoder-decoder self-attention mechanism. Based on aforementioned architecture, models such as MASS \cite{mass}, T5 \cite{t5}, and BART~\cite{bart} have improved quality of the generated text. 

\paragraph{Decoder-only Transformer.} Models such as GPT \cite{gpt2,gpt3} and CTRL \cite{ctrl} employ a single Transformer decoder blocks, which is typically used for language modeling. They apply unidirectional self-attention masking that each token can only attend to previous tokens. 

Besides language modeling, several works also utilize the decoder-only achitecture to generate text conditioned on input text. However, these models do not have an independent module to encode input sequence. Interestingly, they concatenate the input and output sequence with a special token (\eg ``\texttt{[SEP]}'') and employ a novel seq2seq masking~\cite{unilm} that each token in the input sentence can attend to each other and generated tokens can attend to all input tokens and previous generate ones. Compared to unidirectional masking, seq2seq masking is a natural way for decoder-only PLMs to solve conditional  generation tasks, which is similar to the encoder-decoder architecture. \citet{t5} has researched the performance between the above two methods and made a conclusion that the addition of an explicit encoder-decoder attention is beneficial. 

The core of text generation tasks is to learn the semantic mappings from input to output. On one hand, different tasks will correspond to a variety of input data, and we need to develop special techniques to model different data types. On the other hand, the generated text should satisfy important properties in order to cope with different task requirements. Next, we discuss the recent advances with respect to the two aspects, \ie \emph{input} and \emph{output}. 
\section{Modeling Different Data Types from Input}
\label{sec-in}
As discussed in Section~\ref{sec-back}, different text generation tasks usually involve specific kinds of input. In this section, we will introduce three main kinds of input for  text generation, \ie unstructured input, structured input, and multimedia input, and discuss how to model these input data in PLMs.

\subsection{Unstructured Input} 
In NLP research, most of studies focus on modeling unstructured text input (\eg sentence, paragraph, and document). To generate satisfactory output text, it requires an excellent capacity of language understanding beyond surface meaning of individual words in the input text. Thus, \citet{LiuL19} and \citet{ZhengL19} employed PLMs (\eg BERT~\cite{bert}) as text encoder for condensing text into low-dimensional vectors while preserving most of its meaning. Compared with traditional shallow neural models (\eg CNN), PLMs have a large number of parameters encoding massive world knowledge, which is potentially beneficial to capture the core meaning of text. 

In some cases, the input text might be a long document consisting of several sentences and paragraphs. For PLMs trained on sentences or short paragraphs, they are less capable of accurately modeling long-range dependencies in a document. 
Considering this challenge, \citet{ZhangWZ19} and \citet{XuZWWZ20} proposed hierarchical BERT to learn interactions between sentences with self-attention for document encoding. Besides, for capturing inter-sentential relations, DiscoBERT~\cite{XuGCL20} stacked graph convolutional network (GCN) on top of BERT to model structural discourse graphs. By directly operating on the discourse units, DiscoBERT retains capacities to include more concepts or contexts, leading to more concise and informative output text.

We observe that most recent PLMs are pretrained on English text. While, many multilingual generation tasks such as machine translation involve multiple languages and certain languages are low-resource. This challenge hinders the wide application of monolingual PLMs to multilingual text generation tasks. Therefore, \citet{ConneauL19} proposed to learn cross-lingual language models (XLMs) for multilingual language understanding. Based on cross-lingual PLMs, text generation models can still obtain effective input word embeddings even in a low-resource language~\cite{abs-1809-02306}.

\subsection{Structured Input} Structured data (\eg graph and table) is also a critical kind of input for text generation in many real-world applications such as  weather report generation. However, in real-world scenario, it is difficult to collect a large amount of labelled structured data with ground-truth text for training. Since pretrained on large-scale corpus, PLMs encode a large amount of linguistic knowledge  and show excellent few-shot capabilities in many tasks. Motivated by this, \citet{ChenECLW20} and \citet{GongSFQBLL20} explored incorporating PLMs for data-to-text generation, especially in few-shot settings.

When applying PLMs to structured data, a major challenge is how to feed structured data into PLMs, which are originally designed for sequential text. To adapt to the sequential nature of PLMs, \citet{abs-2007-08426} and \citet{MagerANSLFR20} linearized input knowledge graph (KG) and abstract meaning representation (AMR) graph into a sequence of triples, \citet{li2021PLM} introduced an additional graph encoder to encode the input KG, and \citet{GongSFQBLL20} employed a template-based method to serialize input table into text sequence. For example, the attribute-value pair ``\emph{name: jack reynolds}'' will be serialized as a sentence ``\emph{name is jack reynolds}''. However, direct linearization will lose the structural information of original data, which may lead to generating unfaithful text about data. Thus, in addition to generating faithful text, \citet{GongSFQBLL20} proposed an auxiliary reconstruction task for recovering the structural information of input data, which can enhance the capacity of modeling structural information. 

In general, the output text should retain as much as important information from structured data. Therefore, to generate high-fidelity text adhereing to  input, the pointer generator mechanism~\cite{SeeLM17} is adopted to copy words from input knowledge data~\cite{ChenECLW20}. Through grounding PLMs on external knowledge, it is likely to endow a generative model with both rich knowledge and good generalization ability. Besides, \citet{GongSFQBLL20} proposed a content matching loss for measuring the distance between the information in input data and the output text.

\subsection{Multimedia Input} In addition to the above textual data, several attempts have been made to take as input multimedia data (\eg image, video, and speech) such as image caption and speech recognition. Both VideoBERT~\cite{SunMV0S19} and CBT~\cite{abs-1906-05743} conducted pretraining for the video caption task. While, they performed pretraining only for the BERT-based encoder to learn bidirectional joint distributions over sequences of visual and linguistic tokens. So they have to train a separate video-to-text decoder, which tends to cause a \textit{pretrain-finetune discrepancy}. In contrast, Unified VLP~\cite{ZhouPZHCG20} used a shared multi-layer Transformer network for both encoding and decoding. Following UniLM~\cite{unilm}, they pretrained the model on two masked language modeling (MLM) tasks, like cloze tasks designed for sequence-to-sequence LM. Inspired by generative pretraining objectives in GPT, \citet{abs-2003-01473} proposed a cross-modal pretrained model (XGPT) by taking images as inputs and using the image caption task as the basic generative task in the pretraining stage. 

Besides image and video, speech recognition is also hungry for human-transcripted supervised data. So a number of unsupervised and semi-supervised methods are developed to integrate PLMs for weakly-supervised learning. For example, \citet{fan2019unsupervised} proposed an unsupervised approach to pretraining encoder-decoder model with unpaired speech and transcripts. Two pretraining stages are used to extract acoustic and linguistic information with speech and transcripts, which is useful for downstream speech recognition task. 

\section{Satisfying Special Properties for Output Text}
\label{sec-out}
In different text generation tasks, the generated text should satisfy several key  
properties.
In this section, we will introduce three key properties in  text generation, \ie relevance, faithfulness, and order-preservation.

\paratitle{Relevance.} According to the linguistic literatures~\cite{li2021planning}, in text generation, \emph{relevance} refers that the topics in output text is highly related  to the input text. A representative example is the task of dialogue systems, which requires the generated response to be relevant to the input dialogue history.  In addition to the dialogue history, a condition corresponding to the type of response might be also provided as an external input such as the topic of response and the persona of speaker. The generated responses should also be relevant to the condition. Recently, due to the absence of long-term memory, RNN-based models still tend to generate irrelevant output text and lack consistency with input. Therefore, through applying PLMs to the task of dialogue systems, TransferTransfo~\cite{abs-1901-08149} and DialoGPT~\cite{ZhangSGCBGGLD20} were able to generate more relevant and context-consistent responses than traditional RNN-based models. 

Furthermore, to generalize to various types of conditions, \citet{abs-2010-11140} 
	utilized elaborated condition blocks to incorporate external conditions.
	They used BERT for both encoder and decoder by utilizing different input 
	representations and self-attention masks to distinguish the source and 
	target sides of dialogue. On the target (generation) side, a new attention 
	routing mechanism is adopted to generate context-related words. Similar 
	approaches have been used in non-conditioned 
	dialogue~\cite{abs-2006-16779}.

\paratitle{Faithfulness.} Similarly, faithfulness is also a critical property of text, which means the content in generated text should not contradict the facts in input text. Sometimes, it further means the generated text is in accord with the world facts. A representative example is the task of text summarization, which aims to generate faithful text representing the most important information within the original content. Pretrained on large collections of text, PLMs are potentially beneficial to generate faithful text by utilizing background knowledge. \citet{RotheNS20} experimented with a large number of settings to initialize the encoder and decoder with three outstanding PLMs, \ie BERT, GPT and RoBERTa. With pretraining, the models are more aware of the domain characteristics  and less prone to language model vulnerabilities. Consequently, they are more confident in predicting tokens from the document, hence, improving faithfulness.

To improve the level of faithfulness of summary, \citet{KryscinskiPXS18} proposed to decompose the decoder into a contextual network that retrieves relevant parts of the source document and a PLM that incorporates prior knowledge about language generation. Also, to generate faithful text in different target domains, \citet{ted} fine-tuned PLMs on target domains through theme modeling loss. The role of the theme modeling module is to make the generated summary semantically close to the original article.

\paratitle{Order-preservation.} In NLP area, order-preservation denotes that the order of semantic units (word, phrase, etc.) in both input and output text is consistent. The most representative example is the task of machine translation. When translating from source language to target language, keeping the order of phrases consistent in source language and target language will ensure the accuracy of the translation results to some extent. One line of research to achieve the order-preservation property  is to perform semantic alignment in machine translation. \citet{YangHHHJ20} proposed Code-Switching Pre-training (CSP) for machine translation. They extracted the word-pair alignment information from the source and target language, and then applied the extracted alignment information to enhance order-preserving. Besides, it is more common to perform translation across multiple languages, called multilingual machine translation~\cite{ConneauL19}. However, little work can effectively enhance order-preservation for any pairs of languages. Thus, \citet{LinPWQFZL20} proposed mRASP, an approach to pretraining a universal multilingual machine translation model. The key to mRASP is the technique of randomly aligned substitution, which enforces words and phrases with similar meanings across multiple languages to be aligned in the representation space. Also, \citet{abs-1809-02306} focused on aligning word representations of each language, making it possible to preserve the word order consistent cross multiple languages.

\begin{table*}[t]
	\centering
	\begin{tabular}{c|l|l}
		\toprule[1pt]
		
		Data & Categories & Methods \\
		
		\midrule[0.7pt]
		
		\multirow{5}[3]{*}{Input} & Unstructured & \tabincell{l}{BERT acts as text encoders~\cite{LiuL19,ZhengL19}, \\ hierarchical PLMs for document modeling~\cite{ZhangWZ19,XuZWWZ20}, and cross-\\lingual PLMs for multilingual input text~\cite{ConneauL19,abs-1809-02306}.} \\
		\cmidrule{2-3}
		
		& Structured & \tabincell{l}{Linearize KG and AMR graph as triple sequence~\cite{MagerANSLFR20,abs-2007-08426}, \\graph encoder for encoding KG~\cite{li2021PLM}, and serialize table into template-based\\ text sequence~\cite{GongSFQBLL20}.}\\	
		\cmidrule{2-3}
		
		& Multimedia & \tabincell{l}{Video caption~\cite{SunMV0S19,abs-1906-05743}, image caption~\cite{abs-2003-01473}, \\and speech recognition~\cite{fan2019unsupervised}.} \\
		
		\midrule[0.7pt]
		
		\multirow{5}[3]{*}{Output} & Relevance & \tabincell{l}{Fine-tune PLMs in dialogue systems for generating more relevant and context related responses~\\\cite{abs-1901-08149,ZhangSGCBGGLD20}, and generalize to any type of input conditions based on \\BERT~\cite{abs-2010-11140}.} \\
		\cmidrule{2-3}
		
		& Faithfulness & \tabincell{l}{Improve faithfulness with several PLMs~\cite{RotheNS20}, retrieve relevant parts from input \\and incorporate prior knowledge of PLMs~\cite{KryscinskiPXS18}, and generate faithful text in \\different target domains through theme modeling loss~\cite{ted}.} \\
		\cmidrule{2-3}
		
		& \tabincell{l}{Order-\\preservation} & \tabincell{l}{Word-pair alignment~\cite{YangHHHJ20}, a universal multilingual machine translation model~\\\cite{LinPWQFZL20}, and word representation alignment~\cite{abs-1809-02306}.} \\
			
		\bottomrule[1pt]
	\end{tabular}
	\caption{Categories of input types and output properties for text generation.}
	\label{tab:model}
\end{table*}

\section{Fine-tuning Strategies for Text Generation}
\label{sec-finetune}

For text generation with PLMs, a key factor is how to design suitable fine-tuning strategies. In this part, we review several commonly-used fine-tuning strategies from different views. 

\subsection{Data View}
When applying PLMs to text generation tasks especially in a new domain, how to design suitable and effective fine-tuning strategies adapting to the characteristics of new domain is an important consideration.

\paragraph{Few-shot Learning.}
In many text generations, it is difficult and expensive to obtain sufficient annotated data. Owing to the success of pretraining, PLMs can encode massive linguistic and world knowledge, which provides an effective solution to data scarcity. A commonly adopted approach is to plug the existing module with pretrained parameters. Then we fine-tune it with a few, one, or even no examples for the studied task, which are so-called few-shot, one-shot and zero-shot, respectively.

For example in multilingual translation, some low-resource languages lack sufficient parallel corpus. XLM \cite{ConneauL19} proposed to learn cross-lingual language models and can leverage the knowledge learned in high-resource languages to low-resource languages. Using the method proposed in Section~\ref{sec-in}, few-shot learning can also be applied in data-to-text tasks, such as table-to-text generation\cite{ChenECLW20,GongSFQBLL20} and KG-to-text generation\cite{li2021PLM}. \citet{ChenECLW20} directly fed GPT-2 with a small amount of serialized attribute-value pairs and \citet{GongSFQBLL20} further applied multiple tasks to better leverage structured information of tables. Moreover, \citet{li2021PLM} proposed representation alignment to bridge the semantic gap between KG encodings and PLMs in order to enhance the correspondence between KG and text.

\paragraph{Domain Transfer.}
Equipped with vast amounts of parameters and pretrained on large-scale corpus, PLMs have powerful generalization capability. However, they still cannot directly adapt to new domains with large distribution discrepency from pretraining domain~\cite{hendrycks-etal-2020-pretrained}. An effective solution is to continue training PLMs on specific data with pretraining objectives before fine-tuning them on target tasks. Mask prediction is a widely used method, attempting to predict the masked tokens using the remaining ones. There exist several variants of masking ways in domain transfer. \citet{abs-2010-11140} proposed TF-IDF based masking to select more condition-related tokens to mask, in order to focus on domain features. Document masking is usually utilized in summarization task to capture document-level features of long documents \cite{ZhangWZ19}.

\subsection{Task View}
Besides characteristics of new domains, it is also meaningful to consider some special concerns such as language coherence and text fidelity in specific generation tasks when fine-tuning PLMs.

\paragraph{Enhancing Coherence.}
To enhance the language coherence, an important approach is to better model language context during fine-tuning. Models fine-tuned by contrastive learning are good at distinguishing whether a sentence pair is similar or not. Through this method, PLMs are forced to understand the positional or semantic relationship between two sentences, so that they can derive better representations. 

Next sentence prediction (NSP) is a commonly adopted way to judge whether two input sentences are consecutive segments, which can be applied to summarization \cite{ted} and dialog system \cite{abs-1901-08149}. \citet{ZhengL19} proposed to rearrange the sentence order according to their semantic similarities. CBT \cite{abs-1906-05743} proposed noise contrastive estimation (NCE) in cross-modal training to encourage the model to identify the correct video-text pair compared to a set of negative distractors.

Denoising autoencoding (DAE) takes the corrupted text as input and aims to recover the original text. The model fine-tuned with DAE has a strong ability to understand the overall sentences and capture longer-range correlations. For example, TED \cite{ted} utilized DAE to refine essential semantic information for abstractive summarization. XGPT \cite{abs-2003-01473} attempted to model the underlying text-image alignments using image-conditioned denoising autoencoding (IDA), in order to force the model to reconstruct the whole sentence.

\paragraph{Preserving Fidelity.}
Text fidelity refers that how the generated text adheres to the original input information, which is an important aspect to consider in many text generation tasks. The universal structure in PLMs is unable to retain the text fidelity in specific text generation tasks. For the table-to-text generation task,  the structure information of table is required to be encoded. \citet{GongSFQBLL20} proposed to utilize multi-task learning, in order to reconstruct from table embeddings and enforce the match between table embeddings and content embeddings. Besides, the pointer generator \cite{SeeLM17} can be applied to KG-to-text generation to copy the entity and relation information in KG \cite{ChenECLW20}.

\subsection{Model View}
To enhance the quality of generated text, a key is to well train the parameters of PLMs according to task-specific data, so that PLMs can capture the semantic characteristics specially for the generation task. However, as mentioned above, task-specific data is inadequate, thus it is likely to occur the overfitting case when fine-tuned on limited data. In this part, we will introduce several fine-tuning
methods in view of models.

\citet{abs-2004-13835} employed a fixed teacher GPT to preserve the knowledge encoded in another fine-tuned GPT. \citet{distilbert4textgeneration} proposed to utilize a BERT model (teacher) as supervision to guide the Seq2Seq model (student) for better generation performance. Besides, \citet{LiuL19} utilized two optimizers to update the parameters of PLM and initial module separately, in order to solve the discrepancy between two modules. 

There also exist other ways to guide the fine-tuning process. For example, Reinforcement learning can be applied to directly guide models by non-differentiable metrics \cite{ZhangCXW19}, such as ROUGE. \citet{ZhaoWXTZY20} utilized curriculum learning to let the model learn from easy documents to hard documents. Moreover, DIALOGPT \cite{ZhangSGCBGGLD20} implemented a maximum mutual information (MMI) scoring function to alleviate generating bland, uninformative responses.

\section{Conclusion and Future Outlooks}
\label{sec-future}
This paper presents an overview of the recent advances achieved in pretrained language models for text generation. We mainly summarize the extensions of PLMs in modeling  different  data types in input and satisfy special text properties in output. We also  discussed several useful fine-tuning strategies for text generation. 

To advance this field, there are several promising future directions for applying PLMs to text generation.

\paratitle{Model Extension.} Although various extensions have been proposed in Section~\ref{sec-model}, there still exist discrepancies between pretraining and downstream generation tasks. For example, the ``\texttt{[MASK]}'' token in pretraining stage will not be used in fine-tuning stage, which further aggravates the pretraining-finetuning discrepancy. Thus, it further desires to design an appropriate pretraining paradigm for text generation. Moreover, incorporating external knowledge into PLMs during pretraining has been shown to be effective \cite{ernie}, and it is promising to investigate how to inject related knowledge for text generation.


\paratitle{Controllable Generation.} Controllable text generation with PLMs is an interesting direction but still at a very early stage. Controlling some attributes of the generated text has many useful applications such as generating positive response to patients with depression in dialogue systems. However, PLMs are usually pretrained in universal corpus, which is difficult to control the multi-grained attributes of the generated text (\eg sentiment, topic, and coherence). \citet{ctrl} has explored text generation with control codes that govern style, content and task-specific behavior. While, these control codes are preset and coarse-grained. Future work can explore multi-grained control and develop PLMs that are sufficiently steerable.

\paratitle{Model Compression.} Although PLMs with large-scale parameters have achieved success in text generation, these models are challenging to be deployed in resource constrained environments. As a result, it is meaningful to study how to achieve competitive performance with a small number of parameters. Several methods have been proposed to compress PLMs, such as parameter sharing \cite{albert} and knowledge distillation \cite{distilbert}, whereas most of them focused on BERT-based models, and little attention has been paid to compressing PLMs for text generation.

\paratitle{Fine-tuning Exploration.} The direct intention of pretraining is to distill the linguistic knowledge learned in PLMs to downstream generation tasks. And, fine-tuning is the predominant transfer method at present. There could be various ways to transfer knowledge from PLMs to downstream models.  For example, \citet{distilbert4textgeneration} exploited knowledge distillation by adopting BERT as teacher model and a vanilla RNN generation model as student model. Through this method, the linguistic knowledge of BERT can be distilled into the downstream model.


\paratitle{Language-agnostic PLMs.} Nowadays,  almost all the PLMs for text generation are mainly based on English. 
These PLMs will encounter challenges when dealing with non-English generation tasks. 
Therefore, language-agnostic PLMs are worthy to be investigated, which need to capture universal linguistic and semantic features across different languages. An interesting direction is how to reuse existing English-based PLMs for text generation in non-English languages.

\paratitle{Ethical Concern.} Currently, PLMs are pretrained on large-scale corpus crawled from the web without fine-grained filtering, potentially causing  ethical issues such as generating private content about users. Therefore,  researchers should try their best to prevent misusing PLMs. 
For this purpose, we can follow the key steps provided by  \citet{guide4conductrisk}, such as identifying threats and potential impacts and assessing likelihood. Besides, the text generated by PLMs might be prejudiced, which is in line with the bias in training data along the dimensions of gender, race, and religion \cite{gpt3}. Hence, we ought to intervene PLMs for preventing such biases. The research on the general approach is extensive but still preliminary  for PLMs.

\section*{Acknowledgement}
This work was partially supported by the National Key R\&D Program of China under Grant No. 2020AAA0105200, National Natural Science Foundation of China under Grant No. 61872369 and 61832017, Beijing Academy of Artificial Intelligence (BAAI) under Grant No. BAAI2020ZJ0301, Beijing Outstanding Young Scientist Program under Grant No. BJJWZYJH012019100020098, the Fundamental Research Funds for the Central Universities, and the Research Funds of Renmin University of China under Grant No.18XNLG22 and 19XNQ047. Xin Zhao is the corresponding author.


\bibliographystyle{name}
\bibliography{PLM}

\begin{thebibliography}{}

\bibitem[\protect\citeauthoryear{Bahdanau \bgroup \em et al.\egroup
  }{2015}]{BahdanauCB14}
Dzmitry Bahdanau, Kyunghyun Cho, and Yoshua Bengio.
\newblock Neural machine translation by jointly learning to align and
  translate.
\newblock In {\em ICLR}, 2015.

\bibitem[\protect\citeauthoryear{Bao \bgroup \em et al.\egroup
  }{2020}]{abs-2006-16779}
Siqi Bao, Huang He, Fan Wang, Hua Wu, Haifeng Wang, Wenquan Wu, Zhen Guo,
  Zhibin Liu, and Xinchao Xu.
\newblock {PLATO-2:} towards building an open-domain chatbot via curriculum
  learning.
\newblock {\em arXiv preprint arXiv:2006.16779}, 2020.

\bibitem[\protect\citeauthoryear{Blank}{2011}]{guide4conductrisk}
Rebecca~M Blank.
\newblock Guide for conducting risk assessments.
\newblock 2011.

\bibitem[\protect\citeauthoryear{Brown \bgroup \em et al.\egroup }{2020}]{gpt3}
Tom~B. Brown, Benjamin Mann, and Nick~Ryder et~al.
\newblock Language models are few-shot learners.
\newblock In {\em NeurIPS}, 2020.

\bibitem[\protect\citeauthoryear{Chen \bgroup \em et al.\egroup
  }{2020a}]{distilbert4textgeneration}
Yen{-}Chun Chen, Zhe Gan, Yu~Cheng, Jingzhou Liu, and Jingjing Liu.
\newblock Distilling knowledge learned in {BERT} for text generation.
\newblock In {\em ACL}, 2020.

\bibitem[\protect\citeauthoryear{Chen \bgroup \em et al.\egroup
  }{2020b}]{ChenECLW20}
Zhiyu Chen, Harini Eavani, Wenhu Chen, Yinyin Liu, and William~Yang Wang.
\newblock Few-shot {NLG} with pre-trained language model.
\newblock In {\em {ACL}}, 2020.

\bibitem[\protect\citeauthoryear{Conneau and Lample}{2019}]{ConneauL19}
Alexis Conneau and Guillaume Lample.
\newblock Cross-lingual language model pretraining.
\newblock In {\em NeurIPS}, 2019.

\bibitem[\protect\citeauthoryear{Devlin \bgroup \em et al.\egroup
  }{2019}]{bert}
Jacob Devlin, Ming{-}Wei Chang, Kenton Lee, and Kristina Toutanova.
\newblock {BERT:} pre-training of deep bidirectional transformers for language
  understanding.
\newblock In {\em {NAACL-HLT}}, 2019.

\bibitem[\protect\citeauthoryear{Dong \bgroup \em et al.\egroup }{2019}]{unilm}
Li~Dong, Nan Yang, Wenhui Wang, Furu Wei, Xiaodong Liu, Yu~Wang, Jianfeng Gao,
  Ming Zhou, and Hsiao{-}Wuen Hon.
\newblock Unified language model pre-training for natural language
  understanding and generation.
\newblock In {\em NeurIPS}, 2019.

\bibitem[\protect\citeauthoryear{Fan \bgroup \em et al.\egroup
  }{2019}]{fan2019unsupervised}
Zhiyun Fan, Shiyu Zhou, and Bo~Xu.
\newblock Unsupervised pre-training for sequence to sequence speech
  recognition.
\newblock {\em CoRR}, arXiv preprint arXiv:1910.12418, 2019.

\bibitem[\protect\citeauthoryear{Gehring \bgroup \em et al.\egroup
  }{2017}]{GehringAGYD17}
Jonas Gehring, Michael Auli, David Grangier, Denis Yarats, and Yann~N. Dauphin.
\newblock Convolutional sequence to sequence learning.
\newblock In {\em ICML}, 2017.

\bibitem[\protect\citeauthoryear{Gong \bgroup \em et al.\egroup
  }{2020}]{GongSFQBLL20}
Heng Gong, Yawei Sun, Xiaocheng Feng, Bing Qin, Wei Bi, Xiaojiang Liu, and Ting
  Liu.
\newblock Tablegpt: Few-shot table-to-text generation with table structure
  reconstruction and content matching.
\newblock In {\em {COLING}}, 2020.

\bibitem[\protect\citeauthoryear{Gu \bgroup \em et al.\egroup
  }{2020}]{abs-2004-13835}
Jing Gu, Qingyang Wu, Chongruo Wu, Weiyan Shi, and Zhou Yu.
\newblock A tailored pre-training model for task-oriented dialog generation.
\newblock {\em arXiv preprint arXiv:2004.13835}, 2020.

\bibitem[\protect\citeauthoryear{Guan \bgroup \em et al.\egroup
  }{2020}]{guan2020survey}
Wang Guan, Ivan Smetannikov, and Man Tianxing.
\newblock Survey on automatic text summarization and transformer models
  applicability.
\newblock In {\em CCRIS}, 2020.

\bibitem[\protect\citeauthoryear{Hendrycks \bgroup \em et al.\egroup
  }{2020}]{hendrycks-etal-2020-pretrained}
Dan Hendrycks, Xiaoyuan Liu, Eric Wallace, Adam Dziedzic, Rishabh Krishnan, and
  Dawn Song.
\newblock Pretrained transformers improve out-of-distribution robustness.
\newblock In {\em {ACL}}, 2020.

\bibitem[\protect\citeauthoryear{Keskar \bgroup \em et al.\egroup
  }{2019}]{ctrl}
Nitish~Shirish Keskar, Bryan McCann, Lav~R. Varshney, Caiming Xiong, and
  Richard Socher.
\newblock {CTRL:} {A} conditional transformer language model for controllable
  generation.
\newblock {\em arXiv preprint arXiv:1909.05858}, 2019.

\bibitem[\protect\citeauthoryear{Kryscinski \bgroup \em et al.\egroup
  }{2018}]{KryscinskiPXS18}
Wojciech Kryscinski, Romain Paulus, Caiming Xiong, and Richard Socher.
\newblock Improving abstraction in text summarization.
\newblock In {\em EMNLP}, 2018.

\bibitem[\protect\citeauthoryear{Lan \bgroup \em et al.\egroup }{2020}]{albert}
Zhenzhong Lan, Mingda Chen, Sebastian Goodman, Kevin Gimpel, Piyush Sharma, and
  Radu Soricut.
\newblock {ALBERT:} {A} lite {BERT} for self-supervised learning of language
  representations.
\newblock In {\em {ICLR}}, 2020.

\bibitem[\protect\citeauthoryear{Lewis \bgroup \em et al.\egroup }{2020}]{bart}
Mike Lewis, Yinhan Liu, and Naman~Goyal et~al.
\newblock {BART:} denoising sequence-to-sequence pre-training for natural
  language generation, translation, and comprehension.
\newblock In {\em {ACL}}, 2020.

\bibitem[\protect\citeauthoryear{Li \bgroup \em et al.\egroup
  }{2019}]{li2019generating}
Junyi Li, Wayne~Xin Zhao, Ji-Rong Wen, and Yang Song.
\newblock Generating long and informative reviews with aspect-aware
  coarse-to-fine decoding.
\newblock In {\em ACL}, pages 1969--1979, 2019.

\bibitem[\protect\citeauthoryear{Li \bgroup \em et al.\egroup
  }{2020}]{li2020knowledge}
Junyi Li, Siqing Li, Wayne~Xin Zhao, Gaole He, Zhicheng Wei, Nicholas~Jing
  Yuan, and Ji-Rong Wen.
\newblock Knowledge-enhanced personalized review generation with capsule graph
  neural network.
\newblock In {\em CIKM}, pages 735--744, 2020.

\bibitem[\protect\citeauthoryear{Li \bgroup \em et al.\egroup
  }{2021a}]{li2021textbox}
Junyi Li, Tianyi Tang, Gaole He, Jinhao Jiang, Xiaoxuan Hu, Puzhao Xie, Zhipeng
  Chen, Zhuohao Yu, Wayne~Xin Zhao, and Ji-Rong Wen.
\newblock Textbox: A unified, modularized, and extensible framework for text
  generation.
\newblock {\em arXiv preprint arXiv:2101.02046}, 2021.

\bibitem[\protect\citeauthoryear{Li \bgroup \em et al.\egroup
  }{2021b}]{li2021PLM}
Junyi Li, Tianyi Tang, Wayne~Xin Zhao, Zhicheng Wei, Nicholas~Jing Yuan, and
  Ji-Rong Wen.
\newblock Few-shot knowledge graph-to-text generation with pretrained language
  models.
\newblock In {\em Findings of ACL}, 2021.

\bibitem[\protect\citeauthoryear{Li \bgroup \em et al.\egroup
  }{2021c}]{li2021planning}
Junyi Li, Wayne~Xin Zhao, Zhicheng Wei, Nicholas~Jing Yuan, and Ji-Rong Wen.
\newblock Knowledge-based review generation by coherence enhanced text
  planning.
\newblock In {\em SIGIR}, 2021.

\bibitem[\protect\citeauthoryear{Lin \bgroup \em et al.\egroup
  }{2020}]{LinPWQFZL20}
Zehui Lin, Xiao Pan, Mingxuan Wang, Xipeng Qiu, Jiangtao Feng, Hao Zhou, and
  Lei Li.
\newblock Pre-training multilingual neural machine translation by leveraging
  alignment information.
\newblock In {\em {EMNLP}}, 2020.

\bibitem[\protect\citeauthoryear{Liu and Lapata}{2019}]{LiuL19}
Yang Liu and Mirella Lapata.
\newblock Text summarization with pretrained encoders.
\newblock In {\em EMNLP}, 2019.

\bibitem[\protect\citeauthoryear{Mager \bgroup \em et al.\egroup
  }{2020}]{MagerANSLFR20}
Manuel Mager, Ram{\'{o}}n~Fernandez Astudillo, Tahira Naseem, Md.~Arafat
  Sultan, Young{-}Suk Lee, Radu Florian, and Salim Roukos.
\newblock Gpt-too: {A} language-model-first approach for amr-to-text
  generation.
\newblock In {\em ACL}, 2020.

\bibitem[\protect\citeauthoryear{Peters \bgroup \em et al.\egroup
  }{2018}]{elmo}
Matthew~E. Peters, Mark Neumann, Mohit Iyyer, Matt Gardner, Christopher Clark,
  Kenton Lee, and Luke Zettlemoyer.
\newblock Deep contextualized word representations.
\newblock In {\em {NAACL-HLT}}, 2018.

\bibitem[\protect\citeauthoryear{Qiu \bgroup \em et al.\egroup
  }{2020}]{abs-2003-08271}
Xipeng Qiu, Tianxiang Sun, Yige Xu, Yunfan Shao, Ning Dai, and Xuanjing Huang.
\newblock Pre-trained models for natural language processing: {A} survey.
\newblock {\em arXiv preprint arXiv:2003.08271}, 2020.

\bibitem[\protect\citeauthoryear{Radford \bgroup \em et al.\egroup
  }{2019}]{gpt2}
Alec Radford, Jeff Wu, Rewon Child, David Luan, Dario Amodei, and Ilya
  Sutskever.
\newblock Language models are unsupervised multitask learners.
\newblock 2019.

\bibitem[\protect\citeauthoryear{Raffel \bgroup \em et al.\egroup }{2020}]{t5}
Colin Raffel, Noam Shazeer, Adam Roberts, Katherine Lee, Sharan Narang, Michael
  Matena, Yanqi Zhou, Wei Li, and Peter~J. Liu.
\newblock Exploring the limits of transfer learning with a unified text-to-text
  transformer.
\newblock {\em JMLR}, 2020.

\bibitem[\protect\citeauthoryear{Ribeiro \bgroup \em et al.\egroup
  }{2020}]{abs-2007-08426}
Leonardo F.~R. Ribeiro, Martin Schmitt, Hinrich Sch{\"{u}}tze, and Iryna
  Gurevych.
\newblock Investigating pretrained language models for graph-to-text
  generation.
\newblock {\em arXiv preprint arXiv:2007.08426}, 2020.

\bibitem[\protect\citeauthoryear{Rothe \bgroup \em et al.\egroup
  }{2020}]{RotheNS20}
Sascha Rothe, Shashi Narayan, and Aliaksei Severyn.
\newblock Leveraging pre-trained checkpoints for sequence generation tasks.
\newblock {\em TACL}, 2020.

\bibitem[\protect\citeauthoryear{Sanh \bgroup \em et al.\egroup
  }{2019}]{distilbert}
Victor Sanh, Lysandre Debut, Julien Chaumond, and Thomas Wolf.
\newblock Distilbert, a distilled version of {BERT:} smaller, faster, cheaper
  and lighter.
\newblock {\em arXiv preprint arXiv:1910.01108}, 2019.

\bibitem[\protect\citeauthoryear{See \bgroup \em et al.\egroup
  }{2017}]{SeeLM17}
Abigail See, Peter~J. Liu, and Christopher~D. Manning.
\newblock Get to the point: Summarization with pointer-generator networks.
\newblock In {\em ACL}, 2017.

\bibitem[\protect\citeauthoryear{Song \bgroup \em et al.\egroup }{2019}]{mass}
Kaitao Song, Xu~Tan, Tao Qin, Jianfeng Lu, and Tie{-}Yan Liu.
\newblock {MASS:} masked sequence to sequence pre-training for language
  generation.
\newblock In {\em {ICML}}, 2019.

\bibitem[\protect\citeauthoryear{Sun \bgroup \em et al.\egroup
  }{2019a}]{abs-1906-05743}
Chen Sun, Fabien Baradel, Kevin Murphy, and Cordelia Schmid.
\newblock Contrastive bidirectional transformer for temporal representation
  learning.
\newblock {\em arXiv preprint arXiv:1906.05743}, 2019.

\bibitem[\protect\citeauthoryear{Sun \bgroup \em et al.\egroup
  }{2019b}]{SunMV0S19}
Chen Sun, Austin Myers, Carl Vondrick, Kevin Murphy, and Cordelia Schmid.
\newblock Videobert: {A} joint model for video and language representation
  learning.
\newblock In {\em ICCV}, 2019.

\bibitem[\protect\citeauthoryear{Vaswani \bgroup \em et al.\egroup
  }{2017}]{transformer}
Ashish Vaswani, Noam Shazeer, Niki Parmar, Jakob Uszkoreit, Llion Jones,
  Aidan~N. Gomez, Lukasz Kaiser, and Illia Polosukhin.
\newblock Attention is all you need.
\newblock In {\em {NIPS}}, 2017.

\bibitem[\protect\citeauthoryear{Wada and Iwata}{2018}]{abs-1809-02306}
Takashi Wada and Tomoharu Iwata.
\newblock Unsupervised cross-lingual word embedding by multilingual neural
  language models.
\newblock {\em arXiv preprint arXiv:1809.02306}, 2018.

\bibitem[\protect\citeauthoryear{Wolf \bgroup \em et al.\egroup
  }{2019}]{abs-1901-08149}
Thomas Wolf, Victor Sanh, Julien Chaumond, and Clement Delangue.
\newblock Transfertransfo: {A} transfer learning approach for neural network
  based conversational agents.
\newblock {\em arXiv preprint arXiv:1901.08149}, 2019.

\bibitem[\protect\citeauthoryear{Xia \bgroup \em et al.\egroup
  }{2020}]{abs-2003-01473}
Qiaolin Xia, Haoyang Huang, Nan Duan, Dongdong Zhang, Lei Ji, Zhifang Sui,
  Edward Cui, Taroon Bharti, Xin Liu, and Ming Zhou.
\newblock {XGPT:} cross-modal generative pre-training for image captioning.
\newblock {\em arXiv preprint arXiv:2003.01473}, 2020.

\bibitem[\protect\citeauthoryear{Xu \bgroup \em et al.\egroup
  }{2020a}]{XuGCL20}
Jiacheng Xu, Zhe Gan, Yu~Cheng, and Jingjing Liu.
\newblock Discourse-aware neural extractive text summarization.
\newblock In {\em ACL}, 2020.

\bibitem[\protect\citeauthoryear{Xu \bgroup \em et al.\egroup
  }{2020b}]{XuZWWZ20}
Shusheng Xu, Xingxing Zhang, Yi~Wu, Furu Wei, and Ming Zhou.
\newblock Unsupervised extractive summarization by pre-training hierarchical
  transformers.
\newblock In {\em EMNLP}, 2020.

\bibitem[\protect\citeauthoryear{Yang \bgroup \em et al.\egroup
  }{2020a}]{YangHHHJ20}
Zhen Yang, Bojie Hu, Ambyera Han, Shen Huang, and Qi~Ju.
\newblock {CSP:} code-switching pre-training for neural machine translation.
\newblock In {\em EMNLP}, 2020.

\bibitem[\protect\citeauthoryear{Yang \bgroup \em et al.\egroup }{2020b}]{ted}
Ziyi Yang, Chenguang Zhu, Robert Gmyr, Michael Zeng, Xuedong Huang, and Eric
  Darve.
\newblock {TED:} {A} pretrained unsupervised summarization model with theme
  modeling and denoising.
\newblock In {\em {EMNLP} (Findings)}, 2020.

\bibitem[\protect\citeauthoryear{Zaib \bgroup \em et al.\egroup
  }{2020}]{ZaibSZ20}
Munazza Zaib, Quan~Z. Sheng, and Wei~Emma Zhang.
\newblock A short survey of pre-trained language models for conversational
  {AI-A} new age in {NLP}.
\newblock In {\em ACSW}, 2020.

\bibitem[\protect\citeauthoryear{Zeng and Nie}{2020}]{abs-2010-11140}
Yan Zeng and Jian{-}Yun Nie.
\newblock Generalized conditioned dialogue generation based on pre-trained
  language model.
\newblock {\em arXiv preprint arXiv:2010.11140}, 2020.

\bibitem[\protect\citeauthoryear{Zhang \bgroup \em et al.\egroup
  }{2019a}]{ZhangCXW19}
Haoyu Zhang, Jingjing Cai, Jianjun Xu, and Ji~Wang.
\newblock Pretraining-based natural language generation for text summarization.
\newblock In {\em CoNLL}, 2019.

\bibitem[\protect\citeauthoryear{Zhang \bgroup \em et al.\egroup
  }{2019b}]{ZhangWZ19}
Xingxing Zhang, Furu Wei, and Ming Zhou.
\newblock {HIBERT:} document level pre-training of hierarchical bidirectional
  transformers for document summarization.
\newblock In {\em ACL}, 2019.

\bibitem[\protect\citeauthoryear{Zhang \bgroup \em et al.\egroup
  }{2019c}]{ernie}
Zhengyan Zhang, Xu~Han, Zhiyuan Liu, Xin Jiang, Maosong Sun, and Qun Liu.
\newblock {ERNIE:} enhanced language representation with informative entities.
\newblock In {\em {ACL}}, 2019.

\bibitem[\protect\citeauthoryear{Zhang \bgroup \em et al.\egroup
  }{2020}]{ZhangSGCBGGLD20}
Yizhe Zhang, Siqi Sun, Michel Galley, Yen{-}Chun Chen, Chris Brockett, Xiang
  Gao, Jianfeng Gao, Jingjing Liu, and Bill Dolan.
\newblock {DIALOGPT} : Large-scale generative pre-training for conversational
  response generation.
\newblock In {\em ACL}, 2020.

\bibitem[\protect\citeauthoryear{Zhao \bgroup \em et al.\egroup
  }{2020}]{ZhaoWXTZY20}
Xueliang Zhao, Wei Wu, Can Xu, Chongyang Tao, Dongyan Zhao, and Rui Yan.
\newblock Knowledge-grounded dialogue generation with pre-trained language
  models.
\newblock In {\em EMNLP}, 2020.

\bibitem[\protect\citeauthoryear{Zheng and Lapata}{2019}]{ZhengL19}
Hao Zheng and Mirella Lapata.
\newblock Sentence centrality revisited for unsupervised summarization.
\newblock In {\em ACL}, 2019.

\bibitem[\protect\citeauthoryear{Zhou \bgroup \em et al.\egroup
  }{2020}]{ZhouPZHCG20}
Luowei Zhou, Hamid Palangi, Lei Zhang, Houdong Hu, Jason~J. Corso, and Jianfeng
  Gao.
\newblock Unified vision-language pre-training for image captioning and {VQA}.
\newblock In {\em AAAI}, 2020.

\end{thebibliography}

\end{document}